\documentclass[11pt]{article}
\usepackage[margin=1in]{geometry}
\usepackage{authblk} 
\usepackage{hyperref}

\usepackage{amsmath, amssymb}
\usepackage{graphicx}
\usepackage{cite}
\usepackage{color}
\usepackage[most]{tcolorbox}
\usepackage{booktabs}
\usepackage{multirow}
\usepackage{adjustbox}
\usepackage{arydshln}

\begin{document}



\title{MedSyn: Enhancing Diagnostics with Human-AI Collaboration}



\author[1]{Burcu Sayin\thanks{Corresponding author: \texttt{burcu.sayin@unitn.it}}}
\affil[1]{University of Trento, Italy}
\author[2]{Ipek Baris Schlicht}
\affil[2]{Universitat Politècnica de València, Spain}
\author[3]{Ngoc Vo Hong}
\affil[3]{Santa Chiara Hospital, Trento, Italy}
\author[3]{Sara Allievi}
\author[1]{Jacopo Staiano}
\author[4]{Pasquale Minervini}
\affil[4]{The University of Edinburgh, UK}
\author[1]{Andrea Passerini}

\maketitle


\begin{abstract}
Clinical decision-making is inherently complex, often influenced by cognitive biases, incomplete information, and case ambiguity. Large Language Models (LLMs) have shown promise as tools for supporting clinical decision-making, yet their typical one-shot or limited-interaction usage may overlook the complexities of real-world medical practice. In this work, we propose a hybrid human-AI framework, MedSyn, where physicians and LLMs engage in multi-step, interactive dialogues to refine diagnoses and treatment decisions. Unlike static decision-support tools, MedSyn enables dynamic exchanges, allowing physicians to challenge LLM suggestions while the LLM highlights alternative perspectives. Through simulated physician-LLM interactions, we assess the potential of open-source LLMs as physician assistants. Results show open-source LLMs are promising as physician assistants in the real world. Future work will involve real physician interactions to further validate MedSyn’s usefulness in diagnostic accuracy and patient outcomes.
\end{abstract}



\section{Introduction}

In traditional clinical practice, a physician’s diagnosis and treatment plan may be influenced by cognitive biases, incomplete information, or the inherent complexity of the case \cite{Saposnik2016,Ashley2021}. Additionally, physicians often work in time-sensitive, high-pressure environments (e.g., emergency departments), where cognitive overload can increase the risk of misdiagnosis. Recent advancements in 
Large Language Models (LLMs) offer new opportunities for AI-assisted medical decision-making \cite{xie2024llmsdoctorsleveragingmedical,kim2024mdagentsadaptivecollaborationllms,kim2024demonstrationadaptivecollaborationlarge,fan2024aihospital}. We propose that physicians and LLMs can effectively cooperate within multi-step interactive scenarios wherein the LLM's suggestions -- whether accurate or flawed -- serve as opportunities for deeper inquiry and reflection. Thus, in this work, we investigate to what extent such a hybrid cooperative human-AI setup allows physicians to uncover potential oversights, recognize overlooked symptoms, and reconsider treatment options. Unlike static systems that provide one-time recommendations, we propose a dynamic conversational framework that evolves based on real-time interactions, ensuring that physicians maintain control over the clinical decision-making process. Specifically, we explore the collaboration of physicians and LLMs on a specific and sensitive topic: a patient's diagnosis. For instance, if the physician overlooks key symptoms or suggests a suboptimal treatment, the LLM can ask patient-specific follow-up questions or recommend reconsidering the diagnosis. Conversely,  
if an LLM proposes an incorrect diagnosis, the physician can critically examine its reasoning, prompting the model to refine its suggestion. This iterative exchange improves diagnostic accuracy and therapeutic decision-making, serving as a cognitive safety net that aids physicians in complex, ambiguous cases with a higher risk of error. 

This working paper presents our initial efforts on building \textbf{MedSyn}, a medical synergy framework that positions LLMs as conversational partners in clinical decision-making. By fostering human-AI collaboration, MedSyn aims to enhance diagnostics while preserving the physician's critical role in patient care. To evaluate MedSyn, we curate and merge data from MIMIC-IV \cite{MimicIV} and MIMIC-IV-Note \cite{MimicIV-Note,PhysioNet}, creating a diverse set of patient records for model assessment. We then investigate 25 open-source chat-based and medical-domain LLMs to evaluate their capacity for multi-turn engagement. Our analysis highlights both the challenges and opportunities in developing open-source medical dialogue systems. While several models struggled to maintain coherent, multi-turn interactions, others demonstrated the ability to engage in sustained, in-depth discussions about patient conditions. From the 25 evaluated models, we selected three promising candidates for further experimentation—LLaMA3 (8B and 70B) \cite{LlamaModel} and Gemma2 (27B) \cite{GemmaModel}. We also included DeepSeek-R1 \cite{DeepSeekModel}, distilled to Llama3.3-70B-Instruct available via Ollama\footnote{https://ollama.com/library/deepseek-r1:70b}, as a representative of state-of-the-art open-source models that currently fall short in handling complex medical multi-turn dialogues. To assess the role of iterative questioning and collaborative reasoning, we simulate physician–LLM conversations in a controlled setting. Preliminary results show that interactive, multi-step exchanges yield more comprehensive patient assessments and enhance diagnostic clarity. These findings are qualitatively supported by physician analysis of both LLM decisions and their corresponding dialogue traces. As a next step, we aim to replace the simulated physician LLM with real clinicians, enabling direct interaction with the assistant LLM. This will help refine MedSyn for clinical deployment and further validate its utility in real-world medical settings.

\section{MedSyn} \label{sec:MedSyn}

\begin{figure}
    \centering
    \adjustbox{width=.9\linewidth}{
    \includegraphics[width=0.99\linewidth]{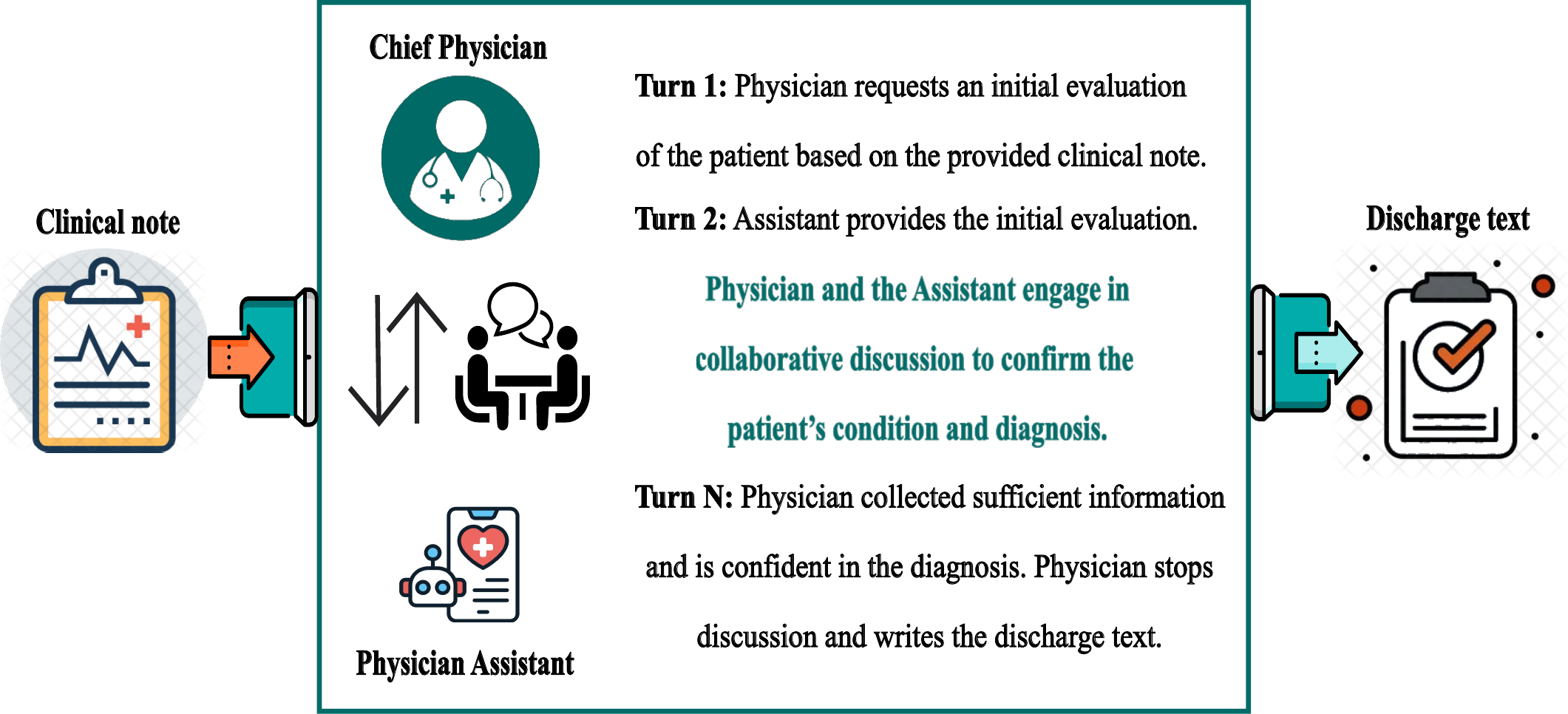}}
    \caption{\textbf{MedSyn} Framework}
    \label{fig:MedSyn}
\end{figure}

Figure \ref{fig:MedSyn} shows the overview of the MedSyn framework. MedSyn receives the clinical note of the patient as input. The clinical note includes several information, including the chief complaint, history of the present illness, physical exam, and pertinent results. Given the limited time physicians have for each patient in the real world, it can be challenging for them to thoroughly review all the details in a clinical note, analyze the patient's condition, and integrate their observations to make an accurate diagnosis. The MedSyn framework supports physicians through an LLM-based virtual assistant that has access to all the details in the clinical note, while the physician is assumed to have access only to the patient's chief complaint. To gather necessary information about the patient and engage in a collaborative discussion, the physician initiates a multi-turn interaction. In the first turn, the physician asks the assistant for an initial evaluation of the patient. In response, the assistant carefully analyzes the clinical note and provides a detailed observation. Following this, the physician and the virtual assistant engage in a dynamic discussion about the patient's condition. This exchange continues until the physician feels they have gathered all the necessary information and is confident in their understanding of the patient’s condition. At this point, the physician concludes the discussion and drafts the discharge text for the patient. The discharge text may include several sections, such as the discharge diagnosis, condition, medications, and the follow-up instructions. For this study, however, we focus solely on the ``diagnosis'' and the corresponding ``ICD-10 codes''\footnote{\href{https://www.icd10data.com/ICD10CM/Codes/}{https://www.icd10data.com/ICD10CM/Codes/}} used by clinicians to code and classify medical diagnoses. 

\section{Experimental Work}

\label{dataset}
We combined MIMIC-IV\footnote{\href{https://physionet.org/content/mimiciv/3.0/}{https://physionet.org/content/mimiciv/3.0/}} \cite{MimicIV,PhysioNet} and MIMIC-IV-Note\footnote{\href{https://physionet.org/content/mimic-iv-note/2.2/}{https://physionet.org/content/mimic-iv-note/2.2/}} \cite{MimicIV-Note} datasets by selecting records with ICD-10 coding~\cite{ICD-10}, which covers diseases from coarse, ``chapter'' level (e.g. E00-E90) to finer granularities (e.g. E10.9 where E10 is a disease category and 10.9 indicates the disease code). The resulting merged dataset contained 122,266 records spanning 5,802 unique diagnoses. Upon analyzing the discharge text field in these records, we observed that most followed a common structure, though certain subsections varied (e.g., the ``major surgical or invasive procedure'' section was present in some records but absent in others). Samples with missing headings or free-form discharge notes hindered effective parsing and prevented the establishment of a standardized format across all records. After consulting with three physicians, we identified the most important sections for our experiments and excluded samples that did not conform to the expected format. Specifically, we selected records that include the following sections in their discharge texts: ``chief complaint, history of present illness, social history, physical exam, pertinent results, major surgical or invasive procedure, brief hospital course, medications on admission, discharge medications, discharge diagnosis, discharge condition, and discharge instructions''. Furthermore, we removed records where the patient’s status was ``deceased'' or ``expired''. This filtering process resulted in a final dataset of 74,850 records. Then, we randomly (seed=13) selected 1,000 records as our test set.
It consists of 2,350 unique diagnoses (on a total of 13,384). The average number of ICD-10 codes appearing in a sample is 5.61. The most common diagnosis is `E78.5' (Hyperlipidemia, Unspecified), while 1,112 diagnoses are identified as the rarest (e.g. `H53.40': Unspecified visual field defects). Since access to this dataset requires completing specialized training, \textit{CITI},\footnote{\href{https://physionet.org/about/citi-course/}{https://physionet.org/about/citi-course/}} we are unable to publicly share our test set and LLM outputs. However, we have detailed our preprocessing steps above and made our code available.\footnote{\href{https://github.com/burcusayin/MedSyn}{See our source code here: https://github.com/burcusayin/MedSyn}}

\subsection{Models \& Frameworks}
We investigated 25 open-source models\footnote{\href{https://huggingface.co/CohereForAI/c4ai-command-r-plus}{command-r-plus:104b}, \href{https://huggingface.co/CohereForAI/c4ai-command-r-v01}{command-r:35b}, \href{https://huggingface.co/openchat/openchat_3.5}{openchat:7b}, \href{https://huggingface.co/mistralai/Mistral-7B-Instruct-v0.2}{mistral:7b}, \href{https://huggingface.co/amazon/MistralLite}{mistrallite:7b}, \href{https://huggingface.co/mistralai/Mixtral-8x7B-Instruct-v0.1}{mixtral:8x7b}, \href{https://huggingface.co/Qwen/Qwen2-7B-Instruct}{qwen2:7b}, \href{https://huggingface.co/epfl-llm/meditron-7b}{meditron:7b}, \href{https://huggingface.co/epfl-llm/meditron-70b}{meditron:70b}, \href{https://huggingface.co/llSourcell/medllama2_7b}{medllama2:7b}, \href{https://huggingface.co/nvidia/Llama3-ChatQA-1.5-70B}{llama3-chatqa:8b and 70b}, \href{https://huggingface.co/collections/meta-llama/meta-llama-3-66214712577ca38149ebb2b6}{llama3:8b and 70b}, \href{https://huggingface.co/meta-llama/Llama-3.1-8B}{llama3.1:8b}, \href{https://huggingface.co/meta-llama/Llama-3.2-3B-Instruct}{llama3.2:3b}, \href{https://huggingface.co/cognitivecomputations/dolphin-2.9-llama3-8b}{dolphin-llama3:8b}, \href{https://huggingface.co/cognitivecomputations/dolphin-2.9-llama3-70b}{dolphin-llama3:70b}, \href{https://huggingface.co/microsoft/Phi-3-mini-4k-instruct-gguf}{phi3:14b}, \href{https://huggingface.co/nvidia/Llama-3.1-Nemotron-70B-Instruct-HF}{nemotron:70b}, \href{https://huggingface.co/lightonai/alfred-40b-1023}{alfred:40b}, \href{https://ollama.com/library/deepseek-r1}{deepseek-R1-Distill-Llama-70B}, \href{https://huggingface.co/collections/allenai/tulu-3-models-673b8e0dc3512e30e7dc54f5}{tulu3:8b and 70b}, \href{https://ollama.com/library/gemma2}{gemma2:27b}} across general-purpose, chat-based, and medical domains, finding that most struggled with multi-turn dialogues. Some chat-based models (e.g., OpenChat:7B \cite{OpenChat}) performed poorly in medical conversations, while certain medical domain models (e.g., Meditron:7B \cite{MeditronPaper} and MedLlama2:7B)\footnote{\href{https://huggingface.co/llSourcell/medllama2\_7b}{https://huggingface.co/llSourcell/medllama2\_7b}} exhibited limitations in handling real-world dialogues. Among the evaluated models, we identified three promising candidates within our experimental setup: Llama3 (8B and 70B) \cite{LlamaModel} and Gemma2:27B \cite{GemmaModel}. To illustrate the challenges even state-of-the-art models face in medical dialogues, we present results with DeepSeek-R1:70B \cite{DeepSeekModel} (Distilled to Llama-70B, available by Ollama\footnote{\href{https://ollama.com/library/deepseek-r1}{https://ollama.com/library/deepseek-r1}}). We implemented our multi-agent environment using Ollama\footnote{\href{https://github.com/ollama/ollama}{https://github.com/ollama}} and Langroid.\footnote{\href{https://github.com/langroid/langroid}{https://github.com/langroid/langroid}}

\subsection{Use cases} \label{sec:cases}

To assess the potential of our framework for real-world deployment in medical decision-making systems, we simulated interactions using LLMs—one serving as the chief physician and another as the physician assistant. As a baseline, we defined the ``phy w/complaint'' scenario, in which the physician LLM receives only the patient’s chief complaint from the clinical note and generates the discharge text without any interaction or dialogue. In contrast, the ``two agent'' setup simulates the collaboration between physicians and assistants in the real world by implementing the MedSyn pipeline (Section~\ref{sec:MedSyn}). Here, the physician agent is limited to the chief complaint, while the assistant agent has access to the complete clinical note, including the history of present illness, physical examination, and pertinent results. Both configurations employ zero-shot prompting, with full prompt details provided below.




\paragraph{Baseline Case}
We use the baseline prompt in the ``phy w/complaint'' case.

\begin{tcolorbox}[fontupper=\small,colback=blue!5!white,colframe=blue!75!black,fonttitle=\small,title=Baseline Prompt]

You are Dr. Ellis, the chief physician responsible for reviewing clinical notes and writing a discharge text for patients. 

**Here is the clinical note for the patient:** $\{$clinicalNote$\}$.

\#\#\# Instructions:
\begin{enumerate}
    \item Carefully analyze the given clinical note to ensure that no symptoms are overlooked.
    \item You are not allowed to ask any questions or make assumptions beyond the information provided in the clinical note.
    \item Once you are ready, write the discharge text for the patient.
    \item The discharge text should include only the `diagnosis' and `codes' fields:
    \begin{itemize}
        \item `diagnosis' field should specify the patient's final diagnosis. Please note that you should decide the final diagnosis.
        \item `codes' field should list the ICD-10 codes corresponding to the diagnosis specified in the `diagnosis' field. Keep in mind that this field is a string, do not use `[]' while listing the codes. 
    \end{itemize}
    \item Remember to refer the clinical note while writing the discharge text. Ensure that the `diagnosis', and `codes' fields are complete and unambiguous; they must not be left empty or unclear.
    \item Return your dischargeText using the TOOL `baseline\_discharge\_text\_tool'. 
\end{enumerate}
\end{tcolorbox}

\paragraph{Two-agent Case}
We use different prompts for the chief physician and physician assistant LLMs.

\begin{tcolorbox}[fontupper=\small,
breakable,pad at break*=0.1mm,colback=blue!5!white,colframe=blue!75!black,fonttitle=\small,title=Chief Physician Prompt]

You are Dr. Ellis, the Chief Physician, collaborating with Dr. Lee, your assistant. 
Your task is to review a clinical note by initiating an evaluation from Dr. Lee and engaging in a natural, focused conversation to assess the patient’s condition. Avoid fabricating interactions or simulating dialogue with Dr. Lee. Instead, clearly articulate your questions or follow-ups, analyze Dr. Lee’s responses, and use this information to guide your decision-making.

Your responsibilities include the following:
\begin{itemize}
    \item Verify the patient’s condition, symptoms, and diagnosis.
    \item Ensure all symptoms are accounted for and thoroughly understand the patient’s condition to deliver optimal care.
    \item Address doubts regarding the diagnosis or treatment plan by conducting further evaluations with Dr. Lee to achieve accurate and effective results.
\end{itemize}

**Here is the clinical note for the patient:** $\{$clinicalNote$\}$.

\#\#\# Instructions:
\begin{enumerate}
    \item Begin by requesting an initial evaluation of the patient from Dr. Lee.
    \item Engage in a collaborative discussion to confirm the patient’s diagnosis. Please note that Dr. Lee has access to a more detailed clinical note, so you MUST consult to Dr. Lee to obtain the necessary information for making the diagnosis.
    \item Keep in mind that you have limited time for every patient. Please avoid duplicate recommendations, conversations, and questions related to treatments. Keep each message CONCISE and to the point.
    \item Once you have gathered sufficient information and are confident in the diagnosis, stop the discussion and write the patient’s discharge text.
    \item The discharge text should include only the `diagnosis' and `codes' fields:
    \begin{itemize}
        \item `diagnosis' field should specify the patient's final diagnosis. Please note that you should decide the final diagnosis.
        \item `codes' field should list the ICD-10 codes corresponding to the diagnosis specified in the `diagnosis' field.
    \end{itemize}
    \item Remember to refer to your discussion with Dr. Lee and the clinical note while writing the discharge text. Ensure that the `diagnosis', and `codes' fields are complete and unambiguous; they must not be left empty or unclear.
    \item Do NOT ask Dr. Lee to check or write your dischargeText. It is YOUR RESPONSIBILITY to write and submit the dischargeText.
    \item Return your dischargeText using the TOOL `discharge\_text\_tool'. Do NOT mention the TOOL `discharge\_text\_tool' to Dr. Lee.
\end{enumerate}
\end{tcolorbox}

\begin{tcolorbox}[fontupper=\small,colback=blue!5!white,colframe=blue!75!black,
  fonttitle=\small,
  title=Physician Assistant Prompt]
You are Dr. Lee, an assistant physician working under the supervision of Dr. Ellis, the chief physician. 
Your role is to review the patient’s clinical notes to perform an initial evaluation, which will support Dr. Ellis in assessing the patient's condition and writing the discharge text. Following your evaluation, you will engage in a collaborative discussion with Dr. Ellis to confirm the diagnosis and determine the next steps.

**Here is the clinical note for the patient:** $\{$clinicalNote$\}$.

\#\#\# Task:
Thoroughly analyze the clinical note and provide a structured summary that includes:
\begin{itemize}
    \item Key symptoms: Highlight notable symptoms that may require further investigation.
    \item Preliminary diagnosis: Offer an initial diagnosis based on the patient’s symptoms and medical history.
    \item Potential complications: Identify any critical issues or risks Dr. Ellis should consider.
    \item Recommendations: Suggest further evaluations if uncertainties remain about the patient’s condition.
\end{itemize}

\#\#\# Instructions:
\begin{enumerate}
    \item Ensure your evaluation is clear, precise, and structured to facilitate an informed discussion. 
    \item In each round of the discussion, limit yourself to a CONCISE message.
    \item Keep in mind that you have limited time for every patient. Please avoid duplicate recommendations, conversations, and questions related to treatments.
\end{enumerate}

\#\#\# Process:
You will first receive a message from Dr. Ellis, asking for your initial assessment. Afterward, you can follow up in each discussion round to collaboratively refine the diagnosis.
\end{tcolorbox}
\section{Results}

Directly comparing discharge texts with LLM responses using standard metrics presents several challenges: (i) Discharge texts lack the conversational tone of LLM responses, (ii) LLMs may generate variable lengths of ICD-10 codes and diagnoses, including occasional hallucinated codes,\footnote{For instance, writing the code M3459 for diagnosis ``Multiple Sclerosis Flare'': the code M3459 does not exists; ``M34'' corresponds to ``systemic sclerosis'' disease which is unrelated to ``multiple sclerosis'' (``G35'').} (iii) Physicians often employ abbreviations and specialized formatting in discharge texts, whereas LLMs produce more standard, conversational sentences, and (iv) The ground truth for diagnoses and ICD-10 codes is longer than LLM outputs. According to two physicians from the in-house annotators, this discrepancy arises because physicians include codes for current and past illnesses based on system recommendations, while LLMs are limited to the information provided in the prompt, which in our case focuses on current symptoms rather than a comprehensive patient history. Thus, specific metrics designed for ICD code detection \cite{Edin2023MedicalCoding} are unsuitable.

\paragraph{ICD-10 Classification}
As stated in \S\ref{dataset}, ICD-10 contains coarse and fine-grained definitions of diseases. In preliminary experiments, we observed that all LLMs tended to not generate fine-grained codes, which could be expected in our zero-shot multi-label classification setup. We explored this issue by discussing several ground truth examples with physicians: they brought to our attention that when selecting ICD-10 subcodes -- often very specific to the diagnosis -- different physicians might choose different codes among those corresponding to the same primary diagnosis; most importantly, it was highlighted how physicians tend to include codes for all the acute or chronic conditions a patient is affected in the patient's medical record, hence including several codes actually unrelated to the specific chief complaint. This characteristic of the ground truth makes the selection of evaluation metrics challenging, as it is impossible to selectively remove the ICD codes unrelated to the chief complaint. For this reason, we resort to compute precision, recall, F1-score, and Jaccard similarity score\footnote{\href{https://github.com/burcusayin/MedSyn/blob/main/src/evaluation/metrics.py}{Please see our code for the evaluation: https://github.com/burcusayin/MedSyn/blob/main/src/evaluation/metrics.py}} on a per-sample basis, and report the mean values in Table~\ref{tab:res:icd}. F1 and Recall show that the agents struggled to accurately predict disease categories, frequently missing ICD codes present in the ground truth. Regarding Precision, all models performed better in predicting disease chapters, a simpler task than detecting disease categories. DeepSeek-R1 and Llama3:70B performed best in the ``phy w/complaint'' case (in terms of precision), with the former excelling in Disease Category and the latter in Disease Chapter. 

\begin{figure*}
    \centering
    \includegraphics[width=.84\linewidth]{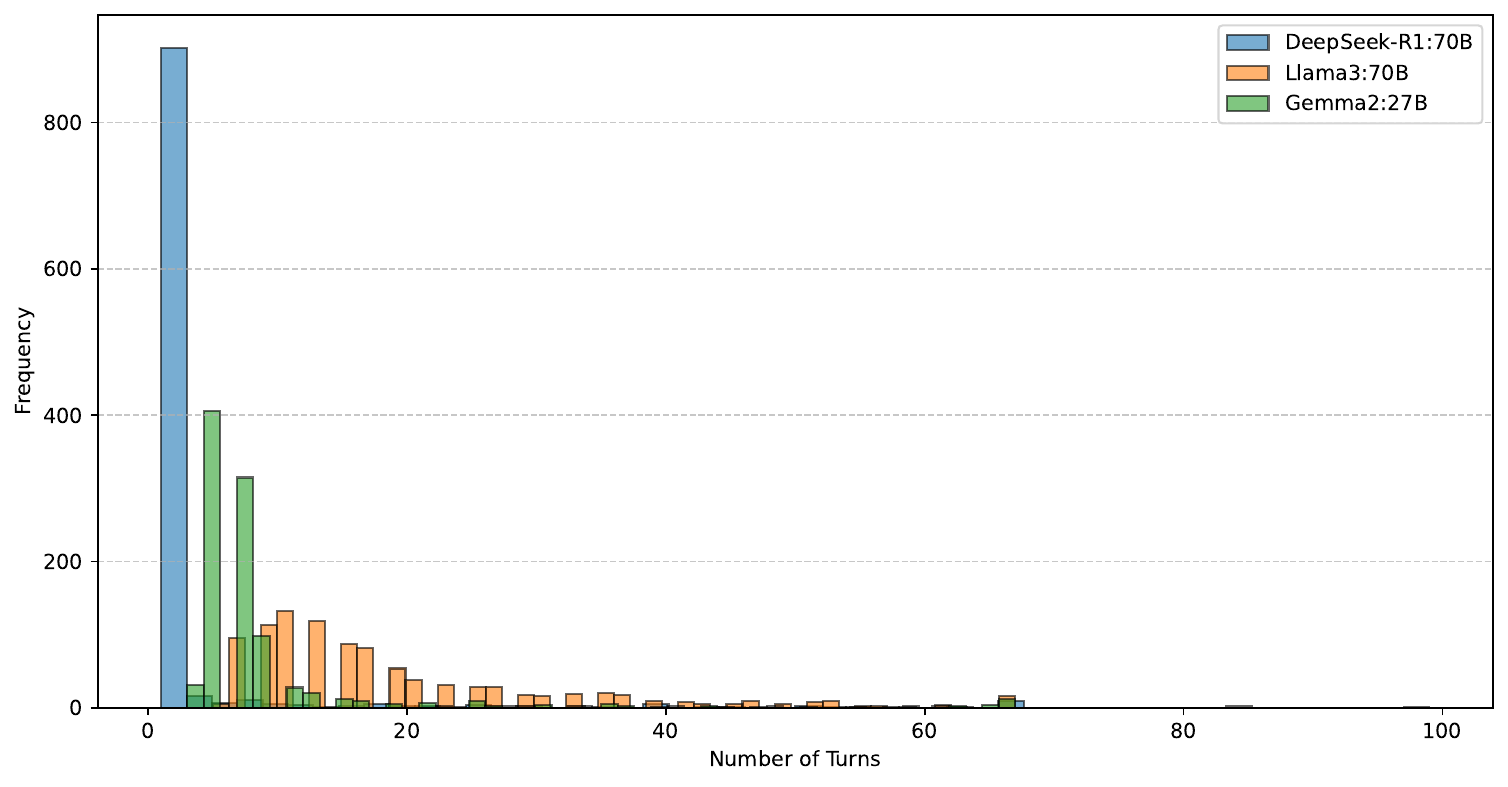}
    \caption{Histogram of multi-turn interactions across physician agents, each engaging with Llama3:8B. DeepSeek-R1 rarely engaged in dialogues. Llama3:70B and Gemma2:27B demonstrated effective interactions.}
    \label{fig:turn_hist}
\end{figure*}

In two-agent case, we observed that DeepSeek-R1 struggled to engage in dialogue. Despite explicitly stating in the prompt that it {\em must} consult the assistant before making a diagnosis, it often relied on internal reasoning and directly generated the discharge text, with minimal interaction with its assistant. Figure~\ref{fig:turn_hist} shows the number of turns each $<$chief physician agent,Llama3:8B$>$ pair produced per sample in the ``two-agent'' case. Notably, DeepSeek-R1:70B engaged in conversations infrequently, whereas Llama:70B exhibited higher interaction, averaging 19.2 turns per sample. Both the Llama3:70B and Gemma2:27B models demonstrated strong performance in engaging in effective dialogues with their assistants and generating well-structured discharge summaries. However, Gemma2:27B was more effective in dialogues, generating the discharge text in 9.3 turns in average. Additionally, Llama3:8B proved to be an effective physician assistant by responding concisely to the chief physician and extracting the necessary information from the clinical note. This is evident from their performance, which closely approaches the performance in ``phy w/full\_note'' case and generates the discharge text without any interaction with the assistant. Our preliminary findings suggest that open-source LLMs hold promise as physician assistants in real-world clinical settings. However, further analysis needed to clarify the limitations and improve performance.

\begin{table*}
    \caption{Performance of Llama3:70B, Gemma2:27B, and DeepSeekR1:70B models as chief physician in ICD-10 disease category and chapter prediction. We use Llama3:8B as the physician assistant. ``phy w/full\_note'' refers to the reference performance of chief physician agents when given full access to the clinical note, as opposed to only the chief complaint, without any interaction with Llama3:8B. We compare the performance in the ``phy w/complaint'' and ``two-agent'' cases, highlighting the best-performing ones.}
    \centering
  \adjustbox{width=.92\linewidth}{
    \begin{tabular}{ll|llll|llll}
         \toprule
         &  & \multicolumn{4}{|c}{\textbf{Disease Category}} & \multicolumn{4}{|c}{\textbf{Disease Chapter}}  \\
         \textbf{Agent} & \textbf{Case} & \textbf{Jaccard} & \textbf{Precision} & \textbf{Recall} & \textbf{F1} & \textbf{Jaccard} & \textbf{Precision} & \textbf{Recall} & \textbf{F1} \\
         \midrule
         Llama3:70B & phy w/complaint & 0.04 & 0.3 & 0.04 & 0.07 & 0.15 & 0.75 & 0.15 & 0.23 \\
         & two-agent & \textbf{0.07} & \textbf{0.37} & \textbf{0.08} & \textbf{0.12} & \textbf{0.22} & \textbf{0.82} & \textbf{0.23} & \textbf{0.34} \\
         \cdashline{2-10}[2pt/2pt]
         & phy w/full\_note & 0.09 & 0.35 & 0.1 & 0.14 & 0.23 & 0.78 & 0.25 & 0.35 \\
         \midrule
         Gemma2:27B & phy w/complaint & 0.03 & 0.26 & 0.04 & \textbf{0.1} & 0.14 & 0.7 & 0.14 & 0.22 \\
         & two-agent & \textbf{0.06}& \textbf{0.38} & \textbf{0.06} & \textbf{0.1} &\textbf{0.18}&\textbf{0.82}&\textbf{0.18}&\textbf{0.28} \\
         \cdashline{2-10}[2pt/2pt]
         & phy w/full\_note & 0.06 & 0.34 & 0.1 & 0.11 & 0.19 & 0.8 & 0.2 & 0.3 \\
         \midrule
         DeepSeekR1:70B & phy w/complaint & \textbf{0.04} & \textbf{0.31} & \textbf{0.04} & \textbf{0.07} & 0.13 & 0.72 & 0.13 & 0.21\\
         & two-agent & \textbf{0.04} & 0.29 & \textbf{0.04} & 0.06 & \textbf{0.14} & \textbf{0.74} & \textbf{0.14} & \textbf{0.23} \\
         \cdashline{2-10}[2pt/2pt]
         & phy w/full\_note & 0.09 & 0.5 & 0.09 & 0.14 & 0.2 & 0.81 & 0.21 & 0.31 \\
         \bottomrule
    \end{tabular}}
    
    \label{tab:res:icd}
\end{table*}

\paragraph{Qualitative  Analysis by Physicians}

The use of LLMs in a healthcare setting has shown interesting results from a clinical perspective. The ``phy w/complaint'' case showed that, starting from the main symptom, LLM was able to identify a possible diagnosis despite having no access to additional clinical and instrumental information. However, it could only align with a subset of the physician's diagnostic hypothesis and was unable to provide a detailed diagnosis. On the other hand, the ``two-agent'' scenario yielded better results in terms of diagnostic precision and completeness. In particular, the Gemma2:27B model made precise diagnoses when interacted with the Llama3:8B model, identifying even rare conditions that could be overlooked by a physician (e.g., Ludwig's angina). The interaction between the physician LLM and the assistant LLM allowed for a more complete diagnosis, as the physician could obtain additional information regarding the patient's characteristics and instrumental exams. In this case, the main challenge was distinguishing between acute and chronic conditions, as there were instances where the chief physician agent identified a pre-existing condition as the primary diagnosis. DeepSeek-R1 did not perform well in ``two-agent'' case, and did not improve the diagnosis compared to ``phy w/complaint'' case, often merely repeating the diagnosis already made. 
Regarding the identification of ICD-10 codes, LLMs were consistently able to identify the general category of the clinical condition, although the specific subcode often differed from the dataset. 
Two-agent scenario is found to be a valuable resource for physicians, as it allows them to interact with an assistant that provides information and often suggests difficult diagnoses. It can be a useful tool in speeding up the diagnostic process.

\section{Related Work}

Prior studies explored multi-LLM frameworks to enhance accuracy and reasoning, primarily focusing on closed-ended questions \cite{Chan2024,Du2023,Jiang2023,Li2023,Liang2024,Liu2023,Sun2024,Wu2023}. However, their applications remain confined to controlled settings, with limited exploration of real-world human-LLM collaboration. Evaluating LLMs’ multi-turn dialogue capabilities is a step toward practical applications. Kwan et al. \cite{Kwan2024} introduced the MT-Eval benchmark, finding that closed-source models outperform open-source ones, though multi-turn dialogues degrade performance due to retrieval difficulties and error propagation. Bai et al. \cite{bai2024MTBench101} proposed MT-Bench-101 to assess LLMs in multi-turn dialogues, noting issues with adaptability and interactivity. Alignment techniques like RLHF \cite{RLHF} and DPO \cite{DPO}, as well as chat-specific designs, offered limited benefits for multi-turn tasks. Campedelli et al. \cite{campedelli2024iwantbreakfree} examined open-source LLMs in goal-driven collaborations and observed mixed success, with models like Mixtral \cite{jiang2024mixtralexperts} and Mistral \cite{jiang2023mistral7b} exhibiting higher failure rates.

In healthcare, LLMs have been explored for clinical note summarization \cite{krishna2021,cai2022}, aiming to assist physicians, though issues such as hallucinations and missing information persist \cite{ben-abacha2023,moramarco2022Human}. Additionally, metrics like ROUGE \cite{ROUGE} and BLEU \cite{BLEU} used to assess summary quality have faced criticism regarding their effectiveness in evaluating clinical content. Furthermore, simulated patient-doctor interactions have been explored to enhance diagnostic accuracy. Liao et al. \cite{liao2023AutoEvalMultiturn} improved accuracy by prompting LLMs to ask clarifying questions, though hallucinations persisted. Liu et al. \cite{Liu2024KDD} introduced the LLM-specific clinical pathway (LCP) to evaluate diagnostic performance using subjective and objective patient data, revealing challenges in handling multi-turn dialogues and clinical specialties, though their study focused solely on the Chinese language. Xie et al. \cite{xie2024llmsdoctorsleveragingmedical} emphasized LLMs as supportive tools rather than replacements, developing the DoctorFLAN dataset and DotaBench to benchmark medical tasks. While most LLMs underperformed, DotaGPT, trained on DoctorFLAN, achieved superior results, demonstrating the dataset's effectiveness. However, its availability only in Chinese limits the generalizability of the findings to other languages. \cite{kim2024mdagentsadaptivecollaborationllms,kim2024demonstrationadaptivecollaborationlarge} proposed MDAgents, a framework that improves LLM effectiveness in complex medical decision-making by dynamically structuring collaboration models. It adapts to clinical needs by assigning LLMs independently or in groups based on task complexity. However, it fails to consider the critical role of physicians in medical decisions. Finally, Fan et al. \cite{fan2024aihospital} proposed the AI Hospital framework for simulated clinical diagnostics, whereas our approach focuses on iterative physician-LLM collaboration to refine clinical reasoning and decision-making.

\section{Conclusion and Future Work}

This work-in-progress paper introduced MedSyn, a dynamic human-AI collaboration framework designed to enhance clinical decision-making through multi-turn, conversational interactions between physicians and LLMs. Unlike traditional, static decision-support tools, MedSyn fosters an iterative diagnostic process where human expertise and AI-generated insights evolve together, aiming to create a safety net in complex medical scenarios. Through controlled simulations and qualitative analysis, we demonstrated that open-source LLMs are promising in meaningfully assisting physicians by uncovering overlooked information, proposing alternative hypotheses, and contributing to more comprehensive diagnostic reasoning. Our results revealed that while model performance varies, open-source LLMs show promise in improving diagnostic completeness and identifying rare conditions. In addition, physician evaluations highlighted the value of AI assistants not only in information retrieval, but also in hypothesis generation and diagnostic refinement. Despite encouraging results, challenges remain in aligning model outputs with clinical standards, particularly in the accurate generation of ICD-10 codes and managing nuances like chronic vs. acute conditions. These findings underscore the importance of continued iteration on evaluation metrics and dialogue strategies.

Future work will involve human-in-the-loop evaluations, enabling real physicians to engage with MedSyn in real-world settings and provide feedback on usability, relevance, and trustworthiness. We also plan to enhance MedSyn’s factual accuracy in clinical reasoning and coding, ensuring more robust and reliable support. This line of research is critical for the responsible integration of AI into clinical workflows—aiming to reduce diagnostic errors, support clinician decision-making, and ultimately improve patient outcomes. MedSyn represents a step toward more adaptive, intelligent healthcare systems where AI serves not as a replacement, but as a reliable and responsive partner in healthcare.

\section*{Acknowledgments}
  Funded by the European Union. Views and opinions expressed are however those of the author(s) only and do not necessarily reflect those of the European Union or the European Health and Digital Executive Agency (HaDEA). Neither the European Union nor the granting authority can be held responsible for them. Grant Agreement no. 101120763 - TANGO. Andrea Passerini also acknowledges the support of the MUR PNRR project FAIR - Future AI Research (PE00000013) funded by the NextGenerationEU.

\section*{Declaration on Generative AI}
During the preparation of this manuscript, the authors utilized ChatGPT and Grammarly to assist with paraphrasing, improving writing style, and refining grammar. After using these tools, the authors reviewed and edited the content as needed and took full responsibility for the publication’s content.


\bibliographystyle{plain}
\bibliography{main}


\end{document}